\documentclass[lettersize,journal]{IEEEtran}
\IEEEoverridecommandlockouts
\usepackage{amsmath,amssymb,amsfonts}
\usepackage{algorithmic}
\usepackage{graphicx}
\usepackage{textcomp}
\usepackage{xcolor}
\usepackage{bm}
\usepackage{color}
 \usepackage{booktabs} 
\usepackage{flushend}
\usepackage{url}
\usepackage{cite}         
\usepackage[colorlinks=true, allcolors=blue]{hyperref}    
\usepackage{soul}

\usepackage[normalem]{ulem}
\usepackage{xcolor}
\usepackage{orcidlink}
\definecolor{darkgreen}{rgb}{0,.4,0}
\definecolor{darkcyan}{rgb}{0,.4,.4}
\newcommand{\REMOVE}[1]%
          {{\color{blue}\sout{#1}}}

\newcommand{\COMMENT}[1]%
          {{\color{darkgreen}\textbf{{Editor: }} {#1}}}

\title{MixerSENet: A Lightweight Framework for Efficient Hyperspectral Image Classification}

\begin{document}

\author{
Mohammed~Q. Alkhatib$^{\orcidlink{0000-0003-4812-614X}}$,~\IEEEmembership{Senior Member,~IEEE,}
Swalpa Kumar Roy$^{\orcidlink{0000-0002-6580-3977}}$,~\IEEEmembership{Senior Member,~IEEE} 
and~Ali Jamali$^{\orcidlink{0000-0002-6073-5493}}$

\thanks{M. Q. Alkhatib is with the College of Engineering and IT, University of Dubai, Dubai, 14143, UAE. (e-mail: mqalkhatib@ieee.org).}
\thanks{S. K. Roy is with the Department of Computer Science and Engineering, Alipurduar Government Engineering and Management College, West Bengal 736206, India (e-mail: swalpa@agemc.ac.in).}
\thanks{A. Jamali is with the Department of Geography, Simon Fraser University, British Columbia 8888, Canada (e-mail: alij@sfu.ca).}}

\maketitle

\begin{abstract}
In this paper, a novel framework, MixerSENet, is introduced for hyperspectral image (HSI) classification, designed to address the challenges of computational efficiency and limited labeled data. The proposed model processes hyperspectral image patches while maintaining consistent size and resolution throughout the network, effectively decoupling the mixing of spatial and channel dimensions. Notably, MixerSENet is lightweight and computationally efficient, requiring fewer parameters compared to traditional models, making it suitable for resource-constrained environments. A squeeze and excitation block is incorporated into the model to refine feature extraction, enhancing the network’s ability to capture more informative features. Experimental results on two benchmark datasets demonstrate that MixerSENet achieves superior performance, reaching an overall accuracy (OA) of 82.47\% on Houston13 dataset and 96.70\% on the Qingyun dataset, outperforming state-of-the-art methods including 3D-CNN, HybridKAN, HSIFormer, SimPoolFormer, and MorphMamba. Furthermore, a detailed analysis of computational efficiency shows that MixerSENet achieves a favorable balance between accuracy and efficiency, with only 53,146 parameters and an low inference time, confirming its practicality for real-world applications. At publication, source code will be publicly available at \url{https://github.com/mqalkhatib/MixerSENet}.
\end{abstract}


\section{Introduction}
Hyperspectral image (HSI) data has been available since the 1980s \cite{qu2022review}. It offers an extensive amount of rich spectral and spatial information, spanning hundreds of narrow contiguous spectral bands from visible to infrared wavelengths. This rich data enables the execution of fine-grained remote sensing tasks that were previously challenging. HSI classification has become a prominent area of research in Remote Sensing (RS), given its wide range of applications in Earth Observation (EO), including land cover and land used mapping and environmental monitoring. The accuracy of HSI classification is a critical factor for the success of these applications. Achieving precise and reliable classification results necessitates the extraction of effective spatial and spectral features.

With the rapid advancements in Deep Learning (DL) technologies, researchers in computer vision have increasingly turned to DL methods, particularly Deep Convolutional Neural Networks (DCNNs), for classifying HSI data. DCNNs have proven to outperform traditional classification methods \cite{alkhatib2023tri, roy2019hybridsn}. In \cite{hu2015deep}, a 1D-CNN method was introduced to extract spectral features. However, relying solely on spectral information is insufficient to achieve high classification accuracy. To address this, \cite{makantasis2015deep} proposed a 2D-CNN approach that incorporates spatial information for HSI classification. Despite this, these methods failed to fully exploit the three-dimensional characteristics of HSI data. To overcome this limitation, \cite{hamida20183} introduced a 3D-CNN approach, which utilizes both spectral and spatial information to further enhance classification performance. Building on this, Roy \textit{et al.} developed the Hybrid Spectral CNN (HybridSN) \cite{roy2019hybridsn}, which combines 2D-CNN and 3D-CNN where the 3D-CNN captures joint spatial–spectral features early in the process, and the 2D-CNN refines the extraction of higher-level spatial features. Numerous other studies have proposed similar approaches \cite{alkhatib2023tri, yu2020simplified}.

Building on the success of transformer models in natural language processing, researchers are increasingly exploring their applications in computer vision and Earth observation \cite{lv2022scvit, cai2022t}. These advanced models have demonstrated significant potential in various fields, including hyperspectral imagery analysis \cite{roy2023multimodal}. However, one major challenge with transformers is their higher requirement for large training datasets compared to Convolutional Neural Networks (CNNs). This makes their application in remote sensing, particularly for hyperspectral imaging with limited labeled data, more difficult. To overcome this challenge, the HSIFormer model \cite{alkhatib2024hsiformer} was introduced. This model employs an efficient Vision Transformer (ViT) with local window attention (LWA) for precise hyperspectral image classification. The aim is to improve classification accuracy while addressing the limitations posed by limited labeled data in remote sensing. However, the major drawback is the significant computational cost and hardware resources required compared to standard CNN classifiers.

In this paper, a novel and lightweight framework, MixerSENet, is presented for HSI classification to address these challenges. The framework processes hyperspectral image patches as input, maintaining consistent size and resolution across the network while effectively decoupling the mixing of spatial and channel dimensions. Notably, MixerSENet is designed to be computationally efficient, requiring fewer parameters compared to conventional models. This design is inspired by the MLPMixer \cite{tolstikhin2021mlp} and PolSARconMixer \cite{jamali2024polsarconvmixer} models, respectively. Additionally, a squeeze and excitation block is incorporated for feature refinement, enhancing the network's ability to capture more informative and robust features.

\begin{figure*}[!t]
\centering
\includegraphics[clip=true, trim = 20 10 20 5,width= 0.9\linewidth]{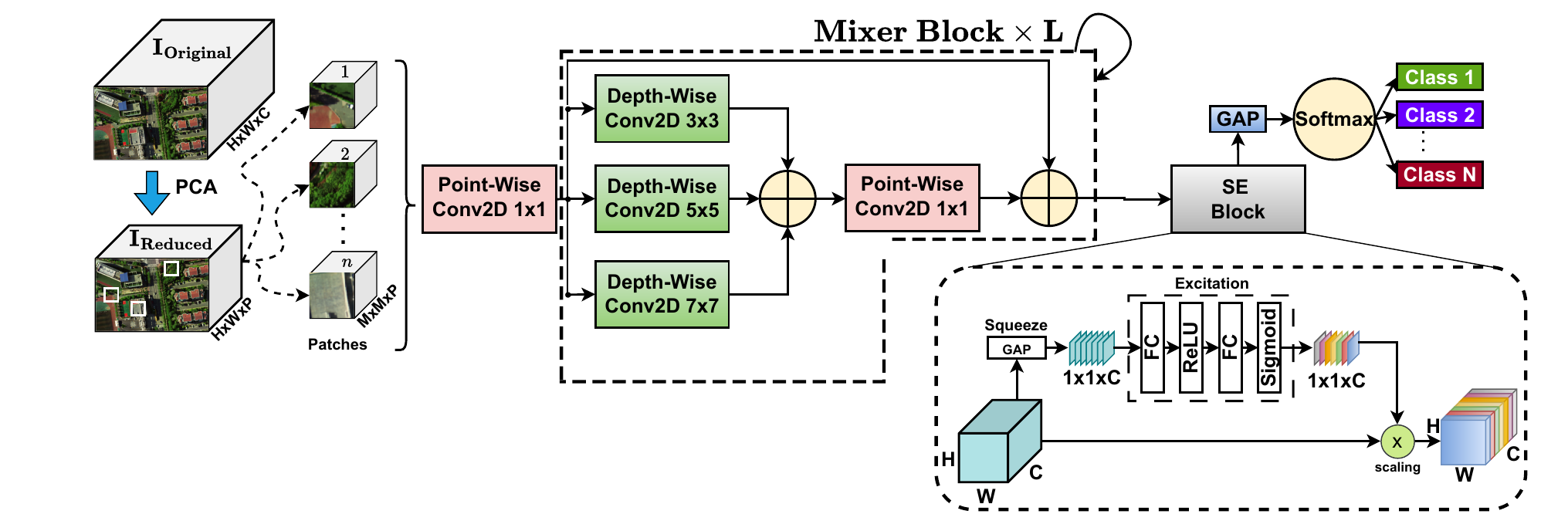}
\vspace{-1em}
\caption{Architecture of the proposed Model}
 \label{fig:model}
\end{figure*}

The rest of the paper is organized as follows: Section \ref{sec:model} explains the architecture and building blocks of the model used in the paper, experimental results and comparisons against state-of-the-art models are discussed in Section \ref{sec:results}, and finally, Section \ref{sec:con} summarizes the paper and states the future direction of this research.

\section{NETWORK ARCHITECTURE} \label{sec:model}
The architecture of the proposed model, shown in Fig. \ref{fig:model}, starts with the input image ($\mathbf{I_{Original}}$) of size ($\mathbf{H \times W \times C}$). Dimensionality reduction is performed using PCA to reduce the number of spectral channels, yielding an output image ($\mathbf{I_{Reduced}}$) with a reduced spectral dimension $\mathbf{P}$, where $\mathbf{P} \ll \mathbf{C}$. The reduced data is then divided into equally sized patches, with each patch processed through point-wise convolution to extract channel features. Depth-wise convolutions with kernel sizes of $3 \times 3$, $5 \times 5$, and $7 \times 7$ are applied to capture spatial features at various scales, followed by a $1 \times 1$ point-wise convolution to mix channel information. This process is repeated $\mathbf{L}$ times to iteratively mix spatial and channel features. A squeeze and excitation (SE) block is introduced for feature refinement, where global average pooling (GAP) compresses the spatial dimensions, followed by fully connected layers with ReLU activation for excitation, and a sigmoid function to calculate the rescaling weights for the input features. Finally, the output is passed through a softmax layer to classify the image patches into one of the predefined classes.

\subsection{Depthwise Convolution}
\begin{figure}[!t]
\centering
\includegraphics[width=0.7\linewidth]{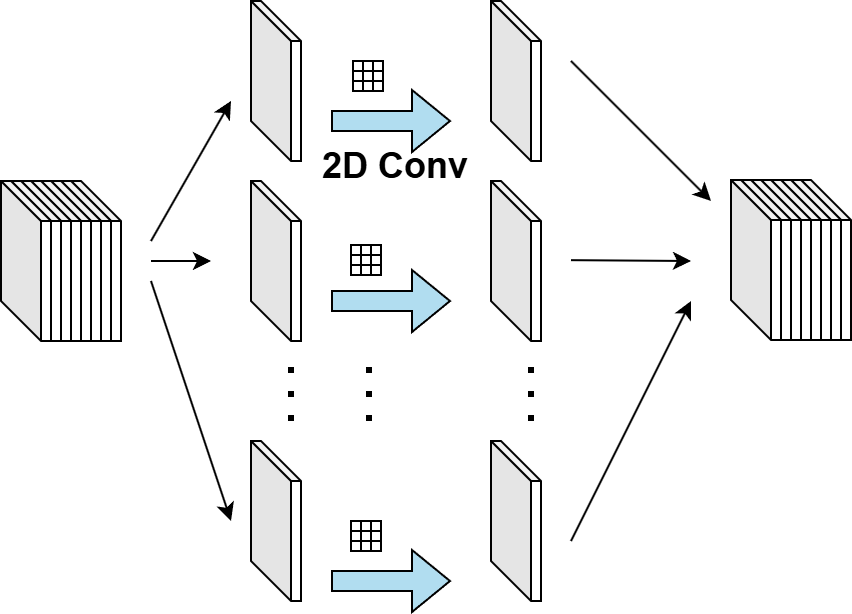}
\vspace{-1em}
\caption{Depthwise Separable Convolution: Input channels are separated, and each is convolved with a spatial filter. The split channels are then concatenated.}
\vspace{-1.25em}
 \label{fig:dwConv}
\end{figure}

Unlike regular 2D convolution, which mixes information across all channels, depthwise convolution applies a separate filter to each channel independently. This greatly reduces parameters and computational cost while preserving the number of channels in the output. By decoupling spatial and channel operations, depthwise convolution extracts channel-specific spatial features efficiently, making it well-suited for resource-constrained environments. The process of depthwise convolution is illustrated in Fig.~\ref{fig:dwConv}.

\subsection{Squeeze and Excitation}
The Squeeze and Excitation (SE) block \cite{zhang2022sem} is designed to improve feature representation by adaptively recalibrating channel-wise responses. This mechanism is particularly beneficial for hyperspectral imagery, where channels correspond to spectral bands that often contain redundant or highly correlated information. By selectively emphasizing more discriminative features while suppressing less informative ones, the SE block enhances the network’s ability to capture meaningful patterns for classification.

Formally, given a transformation $F_{tr}$ that maps the input $X$ to feature maps $u_{c} \in \mathbb{R}^{H \times W \times C}$, where $u_{c}$ is the $c$-th channel, the \textit{squeeze} operation aggregates global spatial information into a compact channel descriptor. This is achieved using global average pooling, which reduces the representation from $H \times W \times C$ to $1 \times 1 \times C$ as:

\begin{equation}
\label{Squeeze}
z_{c} = F_{sq}(u_{c}) = \frac{1}{H \times W} \sum_{i=1}^{H} \sum_{j=1}^{W} u_{c}(i,j).
\end{equation}

\noindent The resulting vector $z$ encodes a global view of channel statistics. The \textit{excitation} step then models channel interdependencies and learns nonlinear relationships between them. It generates a set of weights that highlight the importance of each channel, defined as:

\begin{equation}
\label{excitation}
s = F_{ex}(z, W) = \sigma(W_2 \,\text{ReLU}(W_1 z)),
\end{equation}

\noindent where $\sigma$ denotes the Sigmoid activation, and $W_1$ and $W_2$ are the weights of fully connected layers with a reduction ratio $r$ to control complexity. The resulting channel-wise attention vector $s$ is used to rescale the feature maps $u$, thereby enhancing informative responses while diminishing irrelevant or noisy channels. This recalibration mechanism improves the network’s ability to exploit subtle variations that are critical in hyperspectral image classification.

\section{Experiments and Analysis} 
\label{sec:results}
\begin{figure} [t!]
   \centering
   \includegraphics[clip=true, trim = 20 10 20 5,width= 0.9\linewidth]{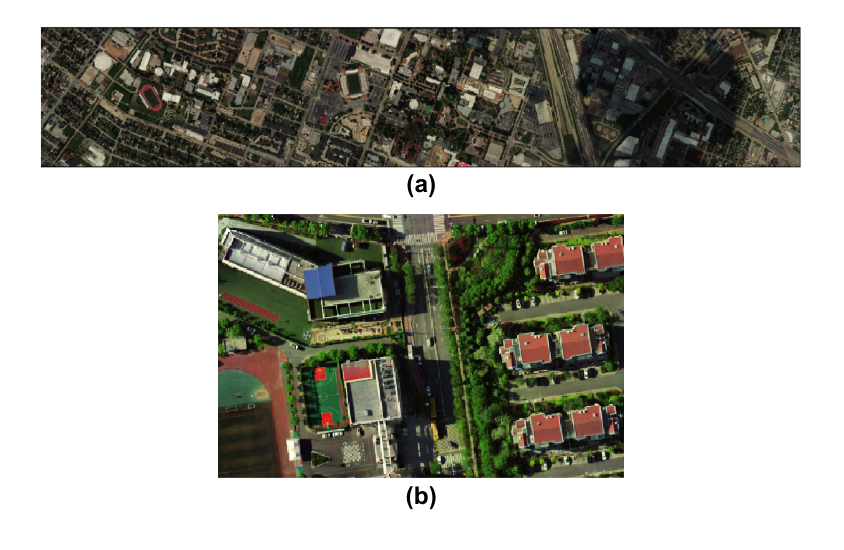}   
   \vspace{-1em}
    \caption{ \label{fig:datasets} RGB Composites: (a) Houston13; (b) QUH-Qingyun.}
\end{figure}

\begin{figure} [t!]
   \centering
   \includegraphics[clip=true, trim = 20 10 20 5,width= 0.9\linewidth]{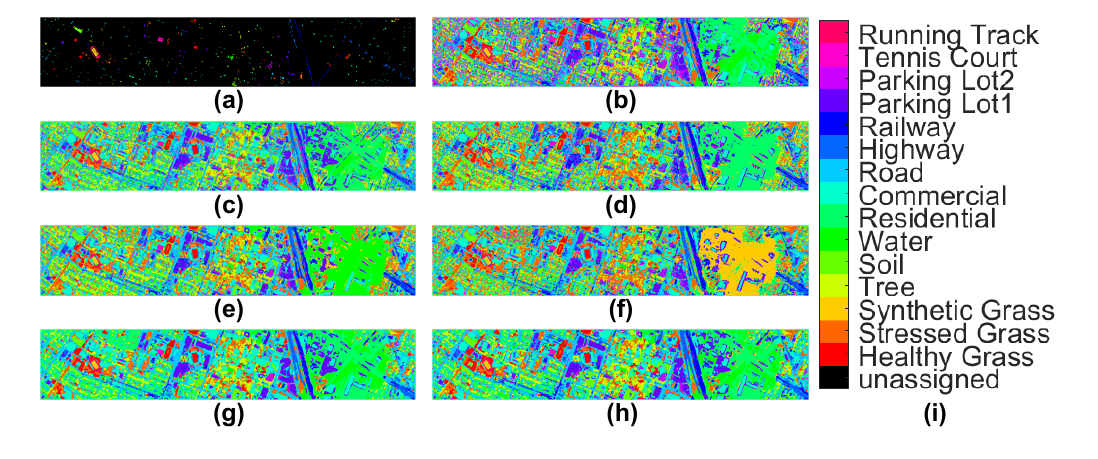}   
   \vspace{-1em}
   \caption{ \label{fig:classification_HO} Classification maps of Houston13 Dataset. (a) Reference Data; (b) 3D-CNN; (c) HybridKAN; (d) HSIFormer; (e) SimPoolFormer; (f) MorphMamba; (g) MixerNet; (h) MixerSENet; (i) Class Labels.}
\end{figure}

\begin{figure} [t!]
   \centering
   \includegraphics[clip=true, trim = 20 10 20 5,width= 0.9\linewidth]{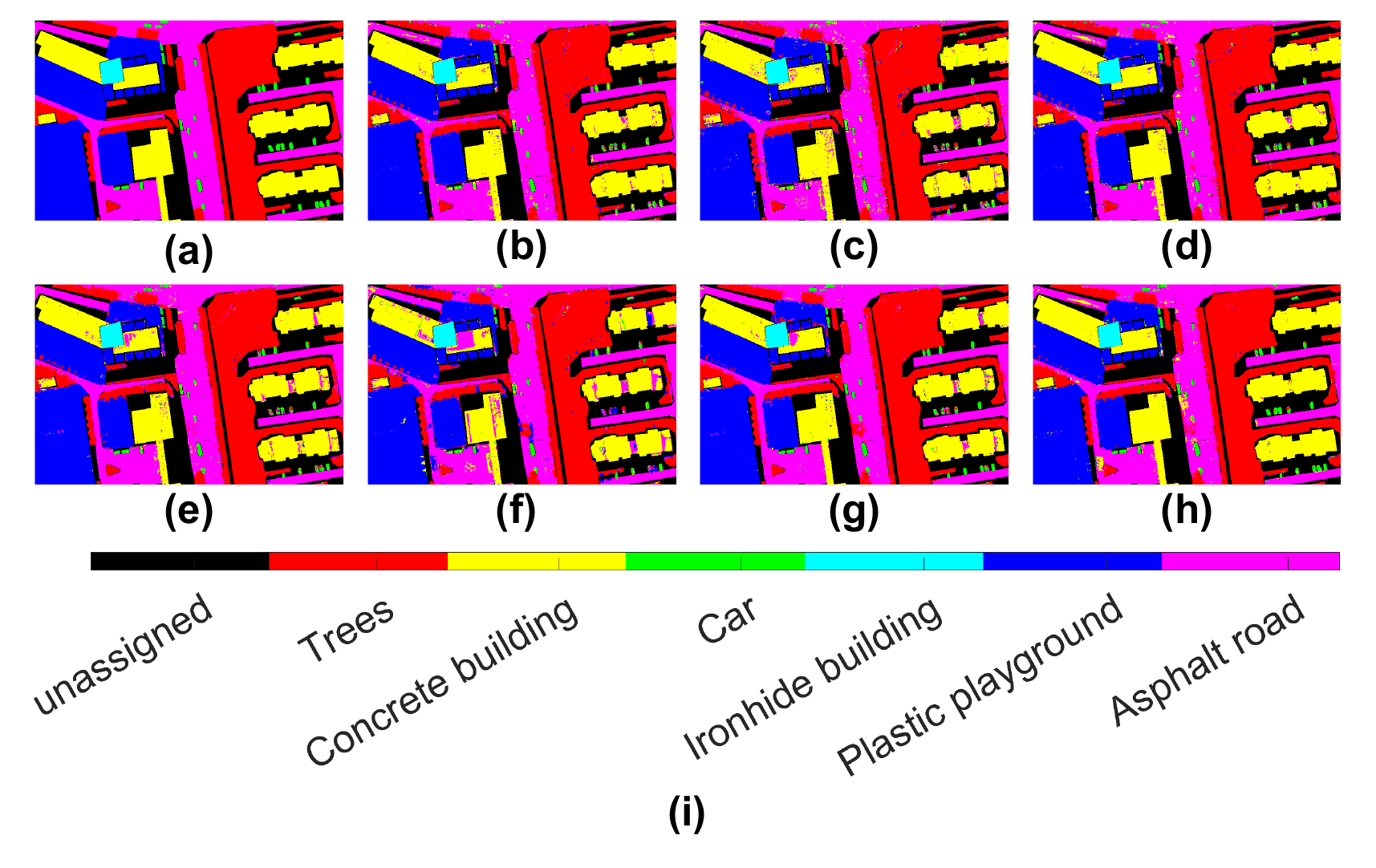}
   \vspace{-1em}
   \caption{ \label{fig:classification_qngn} Classification maps of Qingyun Dataset. (a) Reference Data; (b) 3D-CNN; (c) HybridKAN; (d) HSIFormer; (e) SimPoolFormer;  (f) MorphMamba; (g) MixerNet; (h) MixerSENet; (i) Class Labels.}
\end{figure}

To evaluate the performance of the proposed approach shown in Fig.~\ref{fig:model}, we compare it against several state-of-the-art methods, including 3D-CNN \cite{hamida20183}, the Kolmogorov Arnold Network (HybridKAN) \cite{jamali2024learn}, HSIFormer \cite{alkhatib2024hsiformer}, SimPoolFormer \cite{roy2025simpoolformer} and MorphMamba \cite{ahmad2025spatial}. To assess the impact of attention mechanisms, the model is tested both with and without the attention module. In this context, “MixerNet” denotes the proposed framework without the SE block, whereas “MixerSENet” incorporates the SE block. This distinction effectively serves as an ablation study, isolating the contribution of the SE component to the overall model performance. Performance is measured using Overall Accuracy (OA), Average Accuracy (AA), and the Kappa statistic (Kappa), with the classification accuracy for each class also reported. The evaluation is performed on two widely used hyperspectral datasets: Houston13 and QUH-Qingyun. Fig.~\ref{fig:datasets} shows the RGB composite for both datasets,while Figures \ref{fig:classification_HO}(a) and \ref{fig:classification_qngn}(a) show the available reference class maps. Full description of both datasets is available in \cite{debes2014hyperspectral} and \cite{jamali2024learn}, respectively.

\begin{table}[!t]
\centering
\caption{Classification performance of different methods for the Houston13 dataset.}
\vspace{-1em}
\label{HO_Qualitativeres}
\resizebox{0.95\linewidth}{!}{
\begin{tabular}{ccc|ccccccc}
\hline
\textbf{Class}  & \textbf{\begin{tabular}[c]{@{}c@{}}Train\\ / Val\end{tabular}} & \textbf{Test} & \textbf{\begin{tabular}[c]{@{}c@{}}3D-\\ CNN\end{tabular}} & \textbf{\begin{tabular}[c]{@{}c@{}}Hybrid\\ Kan\end{tabular}} & \textbf{\begin{tabular}[c]{@{}c@{}}HSI-\\ Former\end{tabular}} & \textbf{\begin{tabular}[c]{@{}c@{}}Simpool-\\ Former\end{tabular}} & \textbf{\begin{tabular}[c]{@{}c@{}}Morph-\\ Mamba\end{tabular}} & \textbf{\begin{tabular}[c]{@{}c@{}}Mixer-\\ Net\end{tabular}} & \textbf{\begin{tabular}[c]{@{}c@{}}Mixer-\\ SENet\end{tabular}} \\ \hline
Healthy Grass   & 99                                                             & 1,053         & 81.39                                                      & 80.15                                                         & 80.72                                                          & 80.25                                                              & 81.10                                                           & 81.77                                                         & 81.58                                                           \\
Stressed Grass  & 95                                                             & 1,064         & 85.15                                                      & 76.41                                                         & 85.15                                                          & 85.15                                                              & 83.74                                                           & 81.58                                                         & 83.83                                                           \\
Synthetic Grass & 96                                                             & 505           & 94.46                                                      & 81.19                                                         & 79.80                                                          & 73.07                                                              & 70.69                                                           & 98.42                                                         & 99.80                                                           \\
Tree            & 94                                                             & 1,056         & 92.71                                                      & 88.92                                                         & 88.92                                                          & 90.34                                                              & 83.52                                                           & 91.29                                                         & 89.02                                                           \\
Soil            & 93                                                             & 1,056         & 99.81                                                      & 94.89                                                         & 99.15                                                          & 98.39                                                              & 99.81                                                           & 99.91                                                         & 100.00                                                          \\
Water           & 91                                                             & 143           & 97.90                                                      & 86.71                                                         & 76.92                                                          & 98.60                                                              & 74.13                                                           & 95.80                                                         & 95.10                                                           \\
Residential     & 98                                                             & 1,072         & 91.88                                                      & 79.85                                                         & 90.95                                                          & 72.01                                                              & 83.12                                                           & 88.71                                                         & 94.31                                                           \\
Commercial      & 96                                                             & 1,053         & 69.61                                                      & 63.34                                                         & 66.95                                                          & 68.47                                                              & 69.52                                                           & 73.60                                                         & 78.35                                                           \\
Road            & 97                                                             & 1,059         & 70.07                                                      & 73.65                                                         & 74.22                                                          & 73.18                                                              & 71.29                                                           & 76.02                                                         & 77.43                                                           \\
Highway         & 96                                                             & 1,035         & 55.50                                                      & 42.47                                                         & 44.31                                                          & 49.52                                                              & 40.35                                                           & 38.42                                                         & 44.69                                                           \\
Railway         & 90                                                             & 1,054         & 72.68                                                      & 69.26                                                         & 74.00                                                          & 66.32                                                              & 66.22                                                           & 70.59                                                         & 66.98                                                           \\
Parking Lot1    & 96                                                             & 1,041         & 56.58                                                      & 72.91                                                         & 57.25                                                          & 80.40                                                              & 73.87                                                           & 85.01                                                         & 86.07                                                           \\
Parking Lot2    & 92                                                             & 285           & 92.28                                                      & 83.51                                                         & 86.32                                                          & 87.72                                                              & 72.28                                                           & 86.32                                                         & 82.11                                                           \\
Tennis Court    & 90                                                             & 247           & 100.00                                                     & 76.11                                                         & 100.00                                                         & 100.00                                                             & 93.12                                                           & 98.38                                                         & 100.00                                                          \\
Running Track   & 93                                                             & 473           & 95.35                                                      & 79.49                                                         & 83.30                                                          & 89.43                                                              & 90.91                                                           & 100.00                                                        & 99.37                                                           \\ \hline
\multicolumn{3}{c|}{OA (\%)}                                                                     & \begin{tabular}[c]{@{}c@{}}80.13\\ $\pm$0.52\end{tabular}  & \begin{tabular}[c]{@{}c@{}}75.27\\ $\pm$0.63\end{tabular}     & \begin{tabular}[c]{@{}c@{}}77.38\\ $\pm$0.41\end{tabular}      & \begin{tabular}[c]{@{}c@{}}77.82\\ $\pm$0.46\end{tabular}          & \begin{tabular}[c]{@{}c@{}}76.04\\ $\pm$0.74\end{tabular}       & \begin{tabular}[c]{@{}c@{}}81.23\\ $\pm$0.28\end{tabular}     & \begin{tabular}[c]{@{}c@{}}82.47\\ $\pm$0.25\end{tabular}       \\ \cline{4-10} 
\multicolumn{3}{c|}{AA (\%)}                                                                     & \begin{tabular}[c]{@{}c@{}}83.72\\ $\pm$0.58\end{tabular}  & \begin{tabular}[c]{@{}c@{}}76.64\\ $\pm$0.71\end{tabular}     & \begin{tabular}[c]{@{}c@{}}79.18\\ $\pm$0.46\end{tabular}      & \begin{tabular}[c]{@{}c@{}}80.93\\ $\pm$0.49\end{tabular}          & \begin{tabular}[c]{@{}c@{}}76.87\\ $\pm$0.83\end{tabular}       & \begin{tabular}[c]{@{}c@{}}84.42\\ $\pm$0.37\end{tabular}     & \begin{tabular}[c]{@{}c@{}}85.21\\ $\pm$0.33\end{tabular}       \\ \cline{4-10} 
\multicolumn{3}{c|}{Kappa x 100}                                                                 & \begin{tabular}[c]{@{}c@{}}78.42\\ $\pm$0.39\end{tabular}  & \begin{tabular}[c]{@{}c@{}}73.12\\ $\pm$0.64\end{tabular}     & \begin{tabular}[c]{@{}c@{}}75.48\\ $\pm$0.43\end{tabular}      & \begin{tabular}[c]{@{}c@{}}76.13\\ $\pm$0.48\end{tabular}          & \begin{tabular}[c]{@{}c@{}}74.08\\ $\pm$0.69\end{tabular}       & \begin{tabular}[c]{@{}c@{}}79.63\\ $\pm$0.31\end{tabular}     & \begin{tabular}[c]{@{}c@{}}81.03\\ $\pm$0.27\end{tabular}       \\ \hline
\end{tabular}

}
\end{table}

For the Houston13 dataset, we followed the partitioning scheme published in the 2013 IEEE GRSS Data Fusion Contest. The available training data were evenly divided into training and validation sets, while the provided testing set was kept unchanged. This resulted in approximately 9\% for training, 9\% for validation, and the remaining 82\% for testing. This setup was adopted to ensure consistency with the competition standards. For the Qingyun datasets, the image patches were randomly split into 5\% for training, 5\% for validation, and 90\% for testing and evaluation, the Train/Val/Test splits are shown in Tables \ref{HO_Qualitativeres} and \ref{qngn_Qualitativeres}, it highlights the difference in the amount of samples used for training. A patch size of $9 \times 9$ was used, and the number of principal components was set to 15. To ensure consistency in classification outcomes and minimize the impact of random sample selection, the experiments were repeated 10 times. The final result was determined by averaging the outcomes of these experiments, where recorded values are represented in terms of mean and standard deviation. Additionally, detailed classification results for each category of the best performing iteration were provided. The model was trained for 100 epochs with a batch size of 32. An early stopping strategy was employed, where validation accuracy was evaluated at the end of each epoch. If the validation accuracy did not improve for 10 consecutive epochs, training was halted, and the model was restored to the weights that achieved the highest validation accuracy. The Adam optimizer was used with a learning rate of $1 \times 10^{-3}$. All models were implemented in Python using the Keras framework with TensorFlow as the backend. To ensure a fair comparison, all models were trained under identical conditions.

As shown in Table~\ref{HO_Qualitativeres}, MixerSENet attains the highest overall accuracy (82.47\% $\pm$ 0.25), surpassing MixerNet, 3D-CNN, and transformer-based methods by notable margins, while also yielding the best AA and Kappa values. Class-wise, MixerSENet performs particularly well on Synthetic Grass (99.80\%), Soil (100\%), Residential (94.31\%), Commercial (78.35\%), Road (77.43\%), Parking Lot 1 (86.07\%), and Running Track (99.37\%), indicating strong capability in urban and man-made categories where fine boundaries are critical. Although slightly less competitive for Tree and Water, its performance remains comparable to the strongest baselines. The classification maps in Fig.~\ref{fig:classification_HO} confirm these trends, showing smoother and more continuous labeling in residential and commercial regions when compared against the RGB image, direct comparisons with the reference map remain challenging due to the large proportion of unassigned pixels. Furthermore, the cloudy region observed in the RGB image (Fig.~\ref{fig:datasets}(a)) adds difficulty in distinguishing the classes covered by it, contributing to several misclassifications.

\begin{table}[!t]
\centering
\caption{Classification performance of different methods for the Qingyun dataset.}
\vspace{-1em}
\label{qngn_Qualitativeres}
\resizebox{0.95\linewidth}{!}{
\begin{tabular}{ccc|ccccccc}
\hline
\textbf{Class}     & \textbf{\begin{tabular}[c]{@{}c@{}}Train\\ / Val\end{tabular}} & \textbf{Test} & \textbf{\begin{tabular}[c]{@{}c@{}}3D-\\ CNN\end{tabular}} & \textbf{\begin{tabular}[c]{@{}c@{}}Hybrid-\\ KAN\end{tabular}} & \textbf{\begin{tabular}[c]{@{}c@{}}HSI-\\ Former\end{tabular}} & \textbf{\begin{tabular}[c]{@{}c@{}}SimPool-\\ Former\end{tabular}} & \textbf{\begin{tabular}[c]{@{}c@{}}Morph-\\ Mamba\end{tabular}} & \textbf{\begin{tabular}[c]{@{}c@{}}Mixer-\\ Net\end{tabular}} & \textbf{\begin{tabular}[c]{@{}c@{}}Mixer-\\ SENet\end{tabular}} \\ \hline
Trees              & 13,907                                                         & 250,335       & 96.95                                                      & 95.78                                                          & 95.60                                                          & 97.17                                                              & 95.25                                                           & 98.25                                                         & 97.25                                                           \\
Concrete building  & 8,976                                                          & 161,561       & 96.85                                                      & 93.28                                                          & 96.87                                                          & 93.52                                                              & 84.91                                                           & 98.91                                                         & 93.92                                                           \\
Car                & 689                                                            & 12,405        & 61.53                                                      & 53.59                                                          & 68.03                                                          & 70.91                                                              & 50.80                                                           & 85.63                                                         & 83.32                                                           \\
Ironhide building  & 489                                                            & 8,790         & 98.59                                                      & 96.84                                                          & 96.92                                                          & 97.98                                                              & 98.40                                                           & 99.12                                                         & 99.81                                                           \\
Plastic playground & 10,886                                                         & 195,962       & 97.22                                                      & 95.74                                                          & 97.18                                                          & 96.74                                                              & 96.44                                                           & 97.63                                                         & 96.86                                                           \\
Asphalt road       & 12,797                                                         & 230,351       & 94.74                                                      & 91.37                                                          & 92.92                                                          & 95.88                                                              & 90.29                                                           & 93.19                                                         & 96.06                                                           \\ \hline
\multicolumn{3}{c|}{OA (\%)}                                                                        & \begin{tabular}[c]{@{}c@{}}95.91\\ $\pm$0.46\end{tabular}  & \begin{tabular}[c]{@{}c@{}}93.52\\ $\pm$0.61\end{tabular}      & \begin{tabular}[c]{@{}c@{}}95.10\\ $\pm$0.39\end{tabular}      & \begin{tabular}[c]{@{}c@{}}95.67\\ $\pm$0.44\end{tabular}          & \begin{tabular}[c]{@{}c@{}}91.64\\ $\pm$0.72\end{tabular}       & \begin{tabular}[c]{@{}c@{}}96.04\\ $\pm$0.33\end{tabular}     & \begin{tabular}[c]{@{}c@{}}96.70\\ $\pm$0.21\end{tabular}       \\ \cline{4-10} 
\multicolumn{3}{c|}{AA (\%)}                                                                        & \begin{tabular}[c]{@{}c@{}}90.98\\ $\pm$0.53\end{tabular}  & \begin{tabular}[c]{@{}c@{}}87.77\\ $\pm$0.68\end{tabular}      & \begin{tabular}[c]{@{}c@{}}91.25\\ $\pm$0.41\end{tabular}      & \begin{tabular}[c]{@{}c@{}}92.03\\ $\pm$0.48\end{tabular}          & \begin{tabular}[c]{@{}c@{}}86.02\\ $\pm$0.76\end{tabular}       & \begin{tabular}[c]{@{}c@{}}94.54\\ $\pm$0.36\end{tabular}     & \begin{tabular}[c]{@{}c@{}}95.46\\ $\pm$0.24\end{tabular}       \\ \cline{4-10} 
\multicolumn{3}{c|}{Kappa $\times$ 100}                                                             & \begin{tabular}[c]{@{}c@{}}94.58\\ $\pm$0.44\end{tabular}  & \begin{tabular}[c]{@{}c@{}}91.42\\ $\pm$0.59\end{tabular}      & \begin{tabular}[c]{@{}c@{}}93.51\\ $\pm$0.37\end{tabular}      & \begin{tabular}[c]{@{}c@{}}94.26\\ $\pm$0.42\end{tabular}          & \begin{tabular}[c]{@{}c@{}}88.90\\ $\pm$0.69\end{tabular}       & \begin{tabular}[c]{@{}c@{}}94.76\\ $\pm$0.31\end{tabular}     & \begin{tabular}[c]{@{}c@{}}95.64\\ $\pm$0.20\end{tabular}       \\ \hline
\end{tabular}

}
\end{table}

For the Qingyun dataset, the detailed results are summarized in Table \ref{qngn_Qualitativeres}, which reports the classification performance of different models, while Fig. \ref{fig:classification_qngn} illustrates the corresponding classification maps. In terms of Kappa, MixerSENet achieves 95.64, higher than 3D-CNN (94.58), HybridKAN (91.42), SimPoolFormer (94.26), and HSIFormer (93.51). Although MixerNet obtains a comparable Kappa (94.76), MixerSENet achieves superior OA (96.70\% vs. 96.04\%) and AA (95.46\% vs. 94.54\%), indicating more reliable overall performance. The class-wise results also show MixerSENet performing best in challenging categories such as “Car” (83.32\% compared to 61.53\% for 3D-CNN and 50.80\% for MorphMamba). 

To assess the impact of the depth value $L$, several tests were conducted on the Qingyun dataset while training on 1\% of the data, with results reported in Table~\ref{tab:depth}. Increasing the depth from $\times1$ to $\times5$ improved all metrics, with OA, AA, and Kappa peaking at 93.96\%, 89.45\%, and 92.18\%, respectively, before slightly degrading at $\times6$, suggesting mild overfitting at higher depth. Here, $\times$L denotes the number of repeated Mixer blocks, where each block performs multi-scale spatial mixing with depth-wise convolutions followed by spectral mixing via point-wise convolution. Parameter growth is linear with depth, as each block adds a fixed number of parameters; for instance, x6 has 61,190 parameters compared to 51,270 for x5, an increase of 9,920.

\begin{table}[!t]
\caption{Impact of depth value (L) on classification performance.}
\centering

\vspace{-1em}
\label{tab:depth}
\resizebox{0.9\linewidth}{!}{
\begin{tabular}{ccccccc}
\hline
Depth (L)  & x1     & x2     & x3     & x4     & x5     & x6     \\
\hline
OA         & 92.80  & 93.42  & 93.48  & 93.68  & 93.96  & 93.85  \\
AA         & 86.08  & 87.92  & 88.10  & 88.35  & 89.45  & 89.18  \\
Kappa      & 90.46  & 91.27  & 91.36  & 91.63  & 92.18  & 91.93  \\
\hline
Parameters & 11,590 & 21,510 & 31,430 & 41,350 & 51,270 & 61,190 \\
\hline
\end{tabular}}
\end{table}

\subsubsection{Model Size and Computational Efficiency}

\begin{table}[tb!]
\centering
\caption{Parameters, FLOPs, and MACs of each Model used in the research}
\vspace{-1em}
\label{tab:complexity}
\resizebox{0.9\linewidth}{!}{
\begin{tabular}{lcccc}
\hline
Model        & Parameters & FLOPs ($\times10^6$) & MACs ($\times10^6$) & Inference (m:s) \\ \hline     
3D-CNN        & 397,586        & 0.682   & 0.341&1:37  \\
HybridKAN     & 142,690         & 20.200  & 10.027&03:25 \\
HSIFormer     &  1,373,084        & 18.392  & 9.192&11:00  \\
SimPoolFormer & 771,122        & 57.497  & 28.423 &4:52\\
MorphMamba &  67,650       & 8.592  & 4.294 &7:32 \\
MixerNet      & 52,050         & 7.887   & 3.889&2:30  \\
MixerSENet    & 53,146         & 7.894   & 3.890 &2:32 \\ \hline
\end{tabular}}
\end{table}

Table~\ref{tab:complexity} provides a detailed breakdown  the efficiency of each model in terms of parameters, floating point operations (FLOPs), multiply accumulate operations (MACs), and measured inference time. MixerSENet achieves a balanced trade off with 53,146 parameters, 7.894$\times 10^6$ FLOPs, 3.890$\times 10^6$ MACs, and an inference time of 2 minutes (m) and 32 seconds (s), which is almost identical to MixerNet (2:30) despite the inclusion of the SE block. In contrast, transformer based models are considerably heavier. SimPoolFormer requires 57.497$\times 10^6$ FLOPs and 28.423$\times 10^6$ MACs with an inference time of 4:52, while HSIFormer includes more than 1.3 million parameters and takes 11:00 to complete inference, highlighting the computational burden of transformer architectures. MorphMamba, though lightweight in parameters (67{,}650) and moderate in FLOPs and MACs (8.592 and 4.294$\times 10^6$), still exhibits a long inference time (7:32), consistent with the use of computationally intensive morphological operations and token processing. Other models such as 3D CNN and HybridKAN span the middle range, with 3D CNN showing very low FLOPs and MACs (0.682 and 0.341$\times 10^6$) and the shortest inference time (1:37), yet its accuracy remains lower than MixerSENet. HybridKAN, on the other hand, requires 20.200$\times 10^6$ FLOPs and 10.027$\times 10^6$ MACs and runs in 3:25. Overall, MixerSENet demonstrates strong computational efficiency without compromising accuracy, making it practical for deployment. All inference times were measured on a Windows 10 machine with 64 GB RAM and an NVIDIA GeForce RTX 2080 GPU with 8 GB VRAM.

\subsubsection{Effect of training data}

\begin{figure*}[!t]
\centering
\includegraphics[width=0.9\linewidth]{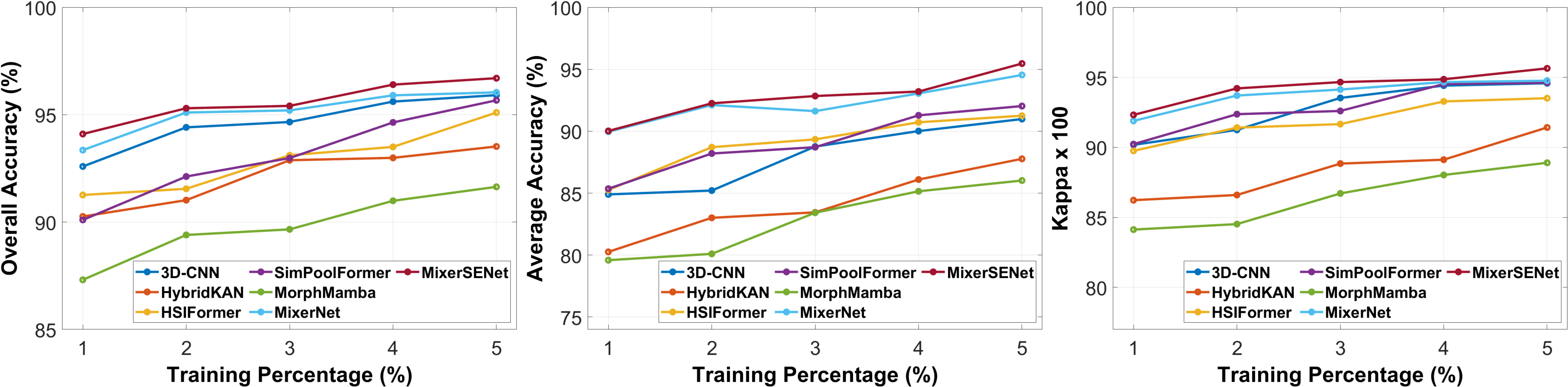}
\vspace{-1em}
\caption{Classification accuracy of Qingyun dataset at different percentages of training data (left) OA (center) AA and (right) Kappa index.}
 \label{fig:qngn_Percentages}
\end{figure*}

To evaluate the impact of training data size on the classification performance of MixerSENet, the model’s performance on the Qingyun benchmark dataset was analyzed across various training ratios, ranging from 1\% to 5\%. The results are presented in Fig.~\ref{fig:qngn_Percentages}. It is observed that MixerSENet achieves a high overall accuracy of 94.10\% with a relatively small amount of training data, outperforming other classifiers such as 3D-CNN (92.59\%), HybridKAN (90.26\%), HSIFormer (91.26\%), SimPoolFormer (90.10\%), MorphMamba (87.31\%), and the Mixer-only network (93.35\%). Notably, a training proportion of 1\% was considered from the reference data. This highlights the effectiveness of MixerSENet in classifying hyperspectral image (HSI) data with limited labeled data, compared to other developed classification algorithms. Furthermore, across all training ratios from 1\% to 5\%, MixerSENet consistently achieves the highest performance in terms of OA, AA, and Kappa. For example, at 5\% training data, MixerSENet obtains 96.70\% OA, 95.46\% AA, and 95.64 Kappa, surpassing MixerNet (96.04 \% OA, 94.54\% AA,and 94.76 Kappa) and significantly outperforming models such as HSIFormer (95.10\% OA, 91.25\% AA, and 93.51 Kappa) and MorphMamba (91.64\% OA, 86.02\% AA, and 88.90 Kappa). These results further confirm not only the robustness of MixerSENet to scarce training data but also its superiority in stability and generalization compared to transformer-based and Mamba-based architectures.

\section{Conclusions}\label{sec:con}
In this paper, MixerSENet, a lightweight and efficient framework for hyperspectral image classification, is introduced. The proposed model utilizes depth-wise convolutions and a squeeze and excitation (SE) block to enhance feature extraction while maintaining computational efficiency. Experimental results demonstrate that MixerSENet outperforms several state-of-the-art models in terms of overall accuracy, average accuracy, and Kappa, with a significant improvement in class-wise performance. Moreover, competitive performance is achieved with fewer parameters, making it an ideal choice for resource-constrained environments. These findings highlight the potential of MixerSENet as a reliable and efficient solution for hyperspectral image classification tasks. 

Future work will focus on further optimizing the model and exploring its applicability to other remote sensing datasets.In particular, while the inclusion of point-wise convolution and the SE block helps mitigate the tendency of depth-wise convolution to struggle with highly correlated spectral bands, this limitation remains and motivates future exploration of alternative spectral mixing strategies. In addition, more sophisticated techniques such as transformers will be considered to further enhance performance in complex classification scenarios.

\vspace{-0.5em}
\bibliographystyle{IEEEtran}
\bibliography{Main_Document}

\end{document}